# Arabic Language Text Classification Using Dependency Syntax-Based Feature Selection

Yannis Haralambous[1], Yassir Elidrissi[1] and Philippe Lenca[1]

*Abstract*— **We study the performance of Arabic text classification combining various techniques: (a) tfidf vs. dependency syntax, for feature selection and weighting; (b) class association rules vs. support vector machines, for classification. The Arabic text is used in two forms: rootified and lightly stemmed. The results we obtain show that lightly stemmed text leads to better performance than rootified text; that class association rules are better suited for small feature sets obtained by dependency syntax constraints; and, finally, that support vector machines are better suited for large feature sets based on morphological feature selection criteria.**

## I. INTRODUCTION

Text classification is an important Natural Language Processing task in the current era of big data, and even more so for Arabic, the fifth language of the world, in terms of numbers of native speakers (according to Wikipedia). It is a supervised classification task, where documents are classified into predefined classes. It can be subdivided in four tasks: pre-processing (where the corpus documents are prepared, text is lemmatized or stemmed, stop-words are removed, morphosyntactic or semantic information is added, etc. so that at the end we have not words but "features"), feature selection (where we select the most relevant features, in order to increase pertinence and decrease memory and CPU cost), feature ranking (where we establish a weight between the features which will be used by the classifier, this step is also called "feature ranking" when the weight is simply the rank of the feature) and finally classification per se.

Feature selection and feature weighting are sometimes merged into a single operation, for example one may use tfidf, which is, strictly speaking, a feature weighting method, for feature selection, by sorting features in decreasing order and taking the most highly ranked. Nevertheless, feature selection and weighting are clearly independent of the classification operation and one can combine feature selection/weighting and classification methods freely, in search of optimal solutions.

### A. Arabic Text Classification

Although quite young, the literature on Arabic text classification is very prolific, starting with the influential works by Sawaf et al. [50] in 2001, El-Kourdi et al. [23] in 2004, Khreisat [37] and Syiam et al. [51] in 2006, El-Halees [21] and Mesleh [42] in 2007, Al-Harbi et al. [2] and Thabtah et al. [54] in 2008. In the last decade, more than seventy papers have been written on the subject. After a peek of publications in 2009 and 2011, the situation is now stable and one observes 4–6 publications per year on the topic.

Let us see, task by task, how the issue of Arabic text classification has evolved over time.

*1) Pre-processing Task:* The task of pre-processing is obviously more difficult for Arabic than for English, because of the complexity of Arabic morphology and the lack of complete vowelization of standard Arabic texts. In the early days, some authors ([23], [19]) have considered using roots, while others ([50], [37], [6], [38]) have evacuated the problem by using $n$-grams of letters as features, instead of words. In the meantime there has been progress on the issue of lemmatization/root extraction/stemming of Arabic ([3], [36], [10], [16], [39], [18], [17], [47], [7], [55], [29], [45]) so that researchers working on Arabic text classification were able to use increasingly efficient stemmers (the best known being the ones by Khoja [36] and by Darwish [16]).

Papers [51], [48], [49], [46] and [12] compare and evaluate various stemming methods. Also [27] attempts an hybrid approach (combining rule-based and statistical stemming), and [35] goes one step farther by replacing words not by roots or stems, but by *concepts* taken from Arabic WordNet [24].

*2) Feature Selection and Weighting Task:* As for these tasks, most authors have used tfidf, sometimes compared to other methods like boolean ranking ("a word belongs to a document or it does not"), tf, log(1 + tf) ([54]) and LTC (a variant of tfidf, cf. [8]). A limited number of authors ([34], [51], [42], [49], [5], [43], [8], [33]) adopt the methodology of distinguishing between feature selection and feature weighting. They use and combine IG (information gain), $\chi^2$, NGL (Ng, Goh & Low), GSS (Galavotti, Sebastiani & Simi), OR (odds ratio) and/or MI (mutual information) for the former, and mostly tfidf for the latter. [56] uses LDA (Latent Dirichlet Allocation) for feature selection (and tfidf for feature weighting). The most synthetic work is [43], which compares 17 different feature selection methods.

Finally, the authors of [8] consider feature selection as a decision problem and attempt an aggregation of feature selection methods (nevertheless, their conclusion is that "combining two or three feature selection methods showed [only] insignificant improvement in classification accuracy").

*3) Classification:* Concerning the last task, namely classification per se, we encounter a great variety of approaches. The four most frequent classifiers used are SVM, Naïve Bayes (as well as its variants, cf. [5] and [33]), kNN and neural networks.

[1]Institut Mines Télécom, Télécom Bretagne, UMR CNRS 6285 Lab-STICC, Université européenne de Bretagne, Technopôle Brest Iroise, CS 83818, 29238 Brest Cedex 3, France `surname.name at telecom-bretagne.eu`

Often several classifiers are evaluated on the same corpus and compared.

Finally, and this brings us closer to our work, a small number of authors ([9], [11], [4], [53], [25], [26]) use rule-based systems, be it decision trees (C4.5, ID3) or class association rules (Apriori, CHARM). The most synthetic papers are [22], [27], [48], [31], [56], which compare various algorithms of quite different nature, on the same corpora.

### B. Our Approach

As far as we can tell, no author uses morphosyntactic information for feature selection, and this is the main contribution of our paper. We show that selecting and weighting features by the syntactic rôle of words, combined with their POS tags is a more efficient feature selection and weighting method than tfidf.

The syntax theory we use is *dependency syntax* [52], [41], [30], and it has the advantage of representing a sentence by a tree entirely built of words of the sentence (i.e., there are no intermediate nodes as in constituency syntax, where we have nodes such as NP "nominal phrase," VP "verbal phrase," etc.). In dependency syntax theory, every sentence contains a "most important word" called *head*, and this is an obvious first candidate when we wish to represent a sentence by a single word/feature.

It turns out, as we will see, that selecting as features the head combined with nouns at distance 1, 2 or 3 of the head is almost as discriminative as selecting the *complete set of nouns of the sentence*. To establish this claim, we evaluate several feature selection strategies, starting by selecting only the heads, then nouns at given distances from the head, etc. We show that for class association rules based on small feature sets, dependency syntax feature selection is significantly more efficient than tfidf.

It should be noted that since dependency syntax operates on the sentence level, we will—as in [30]—restrict ourselves to the sentence level for all calculations. Once classes for all sentences have been predicted, we use a weighted (over the total confidence) majority vote to predict the class of the document. Therefore, tfidf will also be calculated on the sentence level and classifiers (CARs or SVMs) will predict classes for sentences; the document class will be predicted afterwards, as described in [30, Alg. 2]. This explains that our results are slightly inferior to those of [25, Tab. III], which applies the classifier to the entire document at once. By this approach we compare tfidf-based and dependency-based feature selection on a fair level.

This paper is structured as follows: we first describe our corpus and our pre-processing methods (§ II), then the feature selection methods (§ III), the classifiers (§ IV) and the results obtained (§ V). We end the paper by a Conclusion (§ VI).

## II. CORPUS

### A. Selection of Documents

We extracted, 6,000 documents from the *Kalimat* corpus [20]. These documents belong to the following 6 classes (1,000 documents each):

| | words/line | words/doc | lines/doc |
|---|---|---|---|
| Culture | 26.46 | 541.54 | 20.47 |
| Local news (Oman) | 46.91 | 361.73 | 7.71 |
| International news | 27.13 | 499.38 | 18.40 |
| Sports | 35.42 | 325.15 | 9.18 |
| Religion | 39.04 | 482.99 | 12.37 |
| Economy | 33.83 | 457.12 | 13.51 |

The selection of these documents among the 20,291 documents provided by *Kalimat*, has been based on the following criteria:

1) to have short sentences: let $\#p$ be the number of groups of periods, exclamation marks and question marks, divided by the number of words;
2) to have simple sentences, that is sentences with a minimum of connectives tagged WP, WRB WDT (mostly words اللتان, اللي, كيفما, اينما, اللدهان, اللتين, كم, اين, مهما, اللواتي, متى, كلما, اللدهين, لمادها, مادها, من, كيف, الدهى, التى, الدهين, ما, حيث, الدهي, التي): let $\#c$ be the number of connectives divided by the number of words;
3) to have large texts: let $\#w$ be the number of words.

We aggregated these three parameters into a single one:

$$\rho = \frac{\#p \cdot \log(\#w)}{\#c}$$

where we use a logarithm for the number of words criterion in order to limit its strength relatively to the other two criteria. After using $\rho$ to rank documents, we kept the 1,000 first ones for each class.

Finally, the corpus segmentation into sentences has been manually verified, and documents inappropriate for further analysis (poetry, dialectal texts, corrupted documents) have been removed (and replaced by other documents properly ranked).

### B. Word Stemming and Rootification

Having no access to a real Arabic lemmatizer (in the sense of an algorithm replacing verb forms by the 3rd masculine person of the singular present indicative, nouns by the 1st person of the singular nominative and adjectives by the 1st person of the singular nominative masculine), we have chosen to use and compare two extreme approaches: *light stemming* and *rootification*.

*1) Light Stemming:* We have separated clitics from stems using the *Stanford Word Segmenter* [44]. Besides being, in essence, a "very light" stemming method, this step was necessary to be able to use the *Stanford Parser* [28] to obtain dependencies, since it operates solely on segmented text.

Features we obtained by adding to each stem a simplified version of the POS tag given by the *Stanford Parser*: 'n' for nouns, 'np' for proper nouns, 'v' for verbs, and 'x' for all other cases.

*2) Rootification:* Besides raw text, the *Kalimat* corpus includes also morphological information for each word, produced by the *Alkhalil* morphological analyzer [13]. We have kept the root associated to each word (except for proper nouns, that is those having the Stanford tags NNP or DTNNP) and then, to obtain features, combined each root (preceded by the √ symbol, to distinguish a root from a word with exactly the same letters) with a simplified POS tag, as in the case of stemming described above. We call this process *rootification*.

Our corpus *Kalimat* corpus contains 3,689 distinct roots. Among them, 1,441 appear in all classes, 407 in exactly five classes, 325 in exactly four classes, 375 in exactly three classes, 483 in exactly two classes, and 658 in a single class.

In the following table we display the number of roots unique to each class (on the diagonal) and the number of roots appearing in exactly two classes (and appearing more frequently in the class of the line than on the class of the column). The root under the number is the most frequent one for each case.

| | CUL | LOC | INT | SPO | REL | ECO |
|---|---|---|---|---|---|---|
| CUL | **294**<br>**√رغل** | 35<br>√خزف | 60<br>√عبن | 29<br>√تلد | 103<br>√وهج | 16<br>√تبل |
| LOC | 14<br>√شتل | **40**<br>**√زعفر** | 2<br>√هيس | 7<br>√مسر | 9<br>√صيء | 3<br>√وضن |
| INT | 53<br>√سجج | 3<br>√جلف | **60**<br>**√فرقع** | 6<br>√بزز | 13<br>√حرش | 4<br>√حزي |
| SPO | 15<br>√وفض | 3<br>√خوذ | 7<br>√نتر | **50**<br>**√ركل** | 5<br>√جوس | 3<br>√جدف |
| REL | 132<br>√نجس | 15<br>√فسق | 20<br>√زجر | 6<br>√سيغ | **161**<br>**√عبض** | 12<br>√فحش |
| ECO | 21<br>√موه | 8<br>√كشط | 7<br>√هوت | 2<br>√بلت | 5<br>√بطرق | **53**<br>**√لطو** |

For example, there are 161 roots appearing *only* in class Religion and the most frequent one is √عبض; similarly there are 235 (= 132 + 103) roots appearing exactly in classes Religion and Culture, 132 out of which appear more frequently in class Religion and 103 more frequently in class Culture. The more frequent ones in these two cases are resp. √نجس and √وهج.

## III. FEATURE SELECTION AND WEIGHTING METHODS

Two methods are used for feature selection and weighting: *tfidf* and *dependency grammar properties*.

### A. *tfidf*

The legacy feature weighting method tfidf is defined as the the ratio of the frequency of a word in a given document, divided by the number of documents in which it occurs. To first weight and then select the features by tfidf, we take for each sentence $S$ of corpus $\mathbf{C}$ the (stemmed or rootified) words $w \in S$, and apply the following formula:

$$\mathrm{tfidf}_S(w) = \frac{\#\{\tau(w) \in \tau(S)\}}{\#S} \cdot \log\left(\frac{\#\{S \in \mathbf{C}\}}{\#\{S \in \mathbf{C} \mid \tau(w) \in \tau(S)\}}\right)$$

where $\tau$ is either stemming or rootification and # denotes the cardinal of a set (the precise algorithm for calculating tfidf is given in [30, Alg. 4]). Then we rank words according to tfidf value and use the $N$ highest ranked as features for a given sentence.

Here is an example: for the sentence

نشرت اليونان ست بطاريات صواريخ مضادة للصواريخ من طراز باتريوت في مطار تاتوي العسكري شمال اثينا بالقرب من القرية الاولمبية من اجل حماية المجال الجوي للعاصمة ضد اي محاولة اعتداء تستخدم فيها الطائرات خلال اقامة الالعاب الاولمبية من ١٣ الى ٢٩ اغسطس المقبل.

the system has calculated the following features, where n, np and x are our simplified POS tags, √ means that the feature is a root—in which case the third column contains the unrootified word—, and the fourth column contains the tfidf value of the word in the given sentence:

| | | | | | |
|---|---|---|---|---|---|
| n:√صرخ | صواريخ | 0.259 | n:العاصمة | | 0.156 |
| np:تاتوي | | 0.249 | n:اغسطس | | 0.151 |
| x:الاولمبية | | 0.226 | n:√طرز | طراز | 0.150 |
| n:ست | | 0.218 | n:القرب | | 0.140 |
| n:√طري[1] | بطاريات | 0.200 | np:اليونان | | 0.137 |
| n:باتريوت | | 0.198 | x:√جوي | الجوي | 0.134 |
| x:√ضدد | ضد | 0.157 | | | |

The reader may observe that some words (اليونان, الاولمبية, صواريخ, الجوي (المخال)) indeed characterize the sentence, but others (like ست (masculine number six), or القرب without الاولمبية) are rather misleading.

We present in § V-A the results of our text classification algorithm using tfidf as feature selection method.

### B. *Dependency grammar properties*

*Dependency grammar* [52], [41] is a syntactic theory, alternative to the very popular *phrase-structure analysis* [14]. In phrase-structure syntax, trees are built by grouping words into "phrases" (with the use of intermediate nodes NP, VP, etc.), so that the root of the tree represents the entire sentence and its leaves are the actual words. In dependency grammar, trees are built using solely words as nodes (without introducing any additional "abstract" nodes). A single word in every sentence becomes the *head* of the tree. An oriented edge between two words is a *dependency* and is tagged by a representation of some (syntactic, morphological, semantic, prosodic, etc.) relation between the words. For example in the sentence "محمّد يعطي مفتاحًا لفاطمة", the verb "يعطي" is the head of the sentence and we have the following five dependencies:

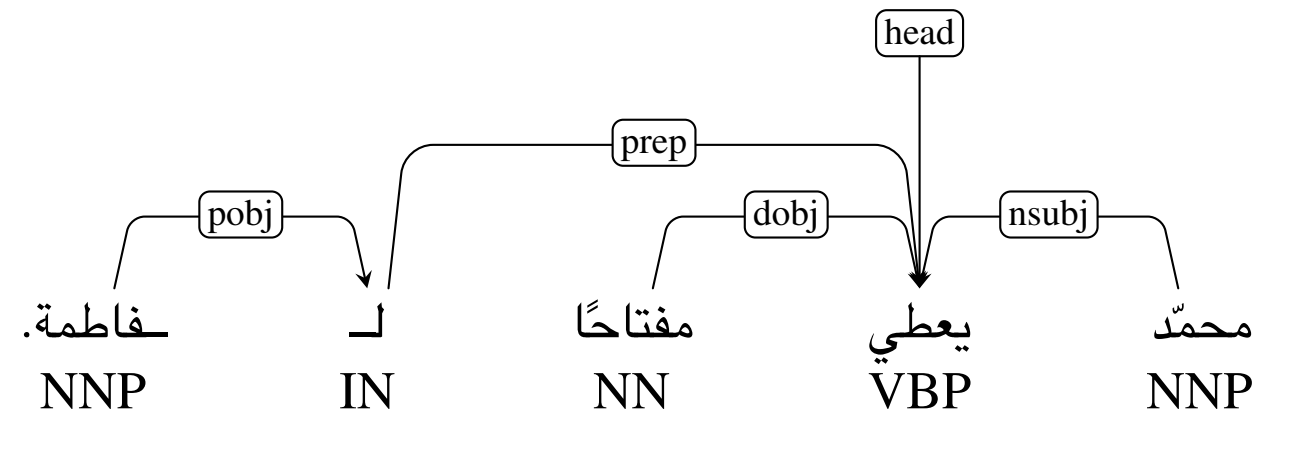

where tags nsubj, dobj, pobj, prep denote "noun subject," "direct object," "prepositional object" and "preposition," and

[1] *Kalimat* gives an erroneous root here: بطاريات being a foreign word, there should be either no root or, if the word is considered as arabified, the pseudo-root √بطر.

NNP, IN, NN, VBP are Stanford POS tags for "proper noun," "preposition," "noun" and "imperfect verb."

To calculate dependencies in our corpus, we used first the *Stanford Word Segmenter* [44] and then the *Stanford Parser* [28] to parse the corpus, and then merged the results with word roots provided by *Kalimat*.

We claim that dependencies provide better features for classification than tfidf, since the latter is based purely on frequencies of words while the former takes advantage of morphosyntactic properties.

To compare with the tfidf method, in Fig. 3, the reader can see the complete dependency tree of the sentence of previous section. Indeed,

- the head of the sentence is √نشر (word نشرت) which has four dependencies:
  - the proper noun DTNN اليونان,
  - the word ست which initiates the following chain of dependencies: ست (NN) ← √طري (word بطاريات) (NNS) ← √صرخ (word صواريخ) (NN) ← √ضدد (word مضادة) (JJ) ← لـ (IN) ← √صرخ (word لصواريخ) (DTNN) which is also the linear order for words,
  - √شمل (word شمال) (NN) which has the dependency اثينا (NNP),
  - and finally, √ضدد (word ضد) (NN) with the dependency √حول (word محاولة) (NN);

In these dependency chains, one finds اليونان, √طري and √صرخ as in the tfidf method, but also اثينا and √حول which seem interesting. It should be noticed that these words have been put forward *without* any frequency-based calculation and solely based on syntactic criteria.

We present in § V-B the results of our text classification algorithm using dependency grammar properties in the feature selection process.

## IV. CLASSIFICATION METHODS

### A. Support Vector Machines

*Support vector machines* (SVMs, cf. [15]) are a nowadays very widespread classification method. They are based on separation of predicted classes by hyperplanes in high dimensional data spaces and provide very good results in general. We use them (in fact, the SVM$^{\text{multiclass}}$ implementation by Thorsten Joachims [32]) on our corpus (rootified or stemmed) with both feature selection methods (tfidf or dependencies).

The disadvantage of SVMs is that they are "black boxes," i.e., it is not possible for the final user to find out why a given result has been obtained, and—besides changing the training corpus and/or the values of some global parameters—one cannot directly modify their behavior.

### B. Class Association Rules

There is another classification method, namely *association rules* [1], which can be used in our context. This method has the advantage that the user can see the rules used for classification and edit them in order to adapt them to a given situation. Here is a formal definition of association rules:

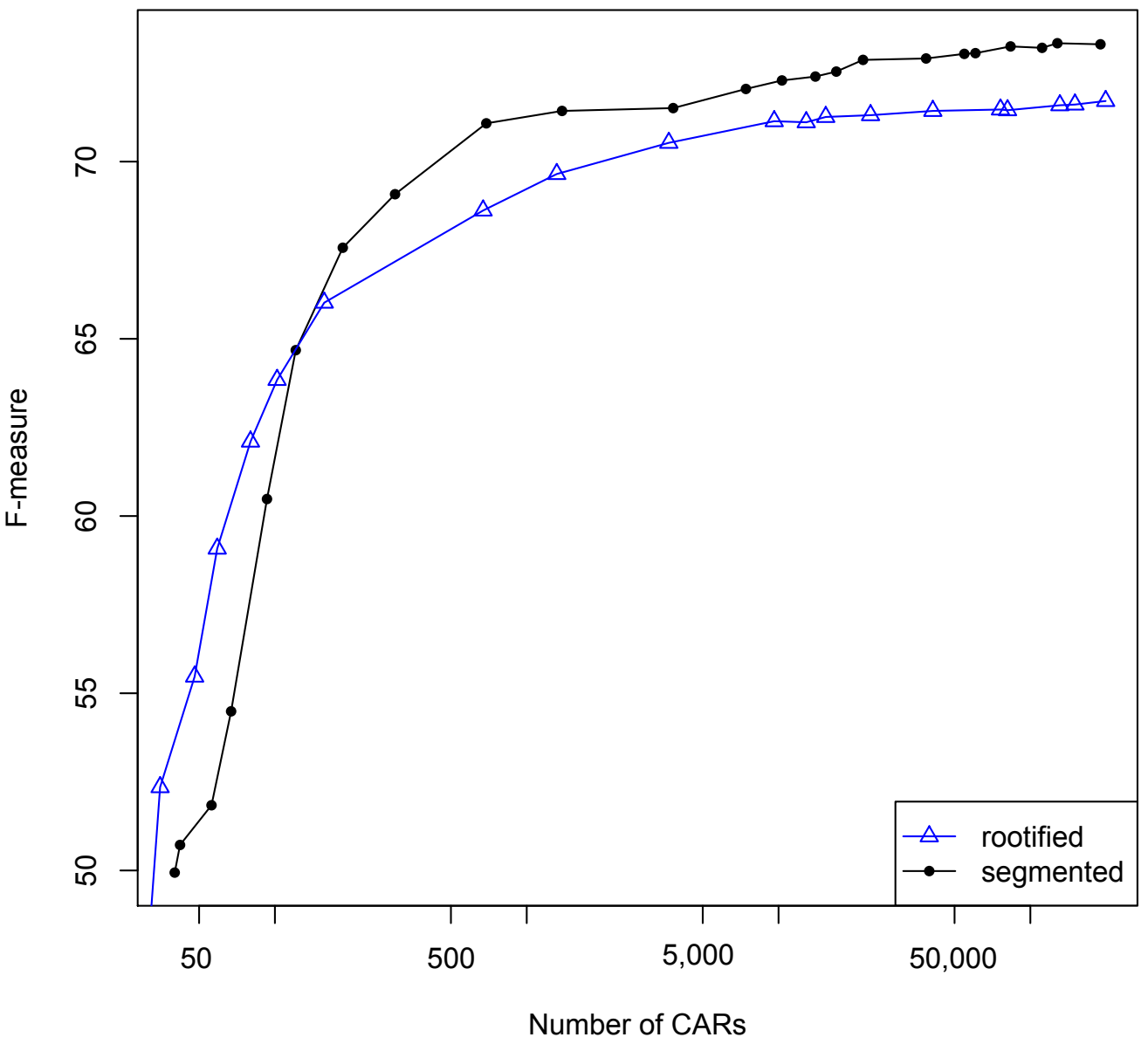

Fig. 1. F-measure as a function of the number of CARs used, when confidence is kept fixed (at 68%).

Let $\mathcal{I}$ be a set of objects called *items* and $\mathcal{C}$ a set of classes. A *transaction* $T$ is a pair $(\{i_1, \ldots, i_n\}, c)$, where $\{i_1, \ldots, i_n\} \subseteq \mathcal{I}$ and $c \in \mathcal{C}$. We denote by $\mathcal{T}$ the set of transactions, by items$(T)$ the set of items (or "itemset") of $T$ and by class$(T)$ the class of $T$.

Let $I$ be an itemset. The *support* of $I$ is defined by

$$\mathrm{supp}(I) := \frac{\#\{T \in \mathcal{T} \,|\, I \subseteq \mathrm{items}(T)\}}{\#\mathcal{T}}.$$

Let $\sigma \in [0, 1]$ be a value called *minimum support*. An itemset $I$ is called *frequent* if its *support* exceeds $\sigma$.

The *confidence* of a transaction $t$ is defined as

$$\mathrm{conf}(t) := \frac{\#\{T \in \mathcal{T} \,|\, \mathrm{items}(t) \subseteq \mathrm{items}(T) \wedge \mathrm{class}(t) = \mathrm{class}(T)\}}{\#\{T \in \mathcal{T} \,|\, \mathrm{items}(t) \subseteq \mathrm{items}(T)\}}.$$

Let $\kappa \in [0, 1]$ be a value called *minimum confidence*. A *class association rule* (or "CAR") $r = (\{i_1, \ldots, i_n\}, c)$ [40] is a transaction with frequent itemset and confidence exceeding $\kappa$.

To classify text with CARs, we consider words as being items, sentences as being itemsets and pairs (sentence, class) as being transactions. The advantage of this technique is that CARs can be easily understood—and hence potentially improved—by the user. Once the classifier is trained, to classify a new sentence we first find all CARs whose items are contained in the sentence, and then use an aggregation technique to choose a predominant class among those of the CARs we found. Finally, from classes predicted for sentences we obtain a class prediction for an entire document, by applying a weighted majority vote technique (cf. [30, Alg. 2]).

We evaluate the classifier by using 10-fold cross validation to obtain average values of recall, precision and F-measure. Comparing rule-based classification methods is tricky because one can always increase F-measure performance by increasing the number of rules, which results in overfitting. To avoid this phenomenon and compare methods in a fair way, we have chosen to fix the number $N_R$ of rules and find values of minimal support and confidence so that F-measure is maximal under this constraint. (See [30, § 2] for more details on the algorithms used.) In Fig. 1 the reader can see F-measure as a function of the number of CARs, obtained by varying support while keeping confidence fixed. At $N_R$ = 10,000, the CPU charge is not yet very important and F-measure remains relatively stable (in this example, increasing $N_R$ from 10,000 to 200,000, increases F-measure by only 0.55%, while processing time is multiplied by a factor of 17!). This is why we have chosen to fix the number of rules to $N_R$ = 10,000 for all subsequent calculations.

Here is an example of the thirty first (by order of decreasing support) CARs we get by using dependency grammar properties:

| | | | | |
|---|---|---|---|---|
| 1 | religion | ← | n:فتاوى n:احكام | 0.173% |
| 2 | culture | ← | n:الفيلم | 0.145% |
| 3 | economy | ← | n:اوبك | 0.142% |
| 4 | economy | ← | n:√بوس n:√سهم | 0.141% |
| 5 | sports | ← | n:منتخب | 0.141% |
| 6 | sports | ← | n:المباراة | 0.135% |
| 7 | economy | ← | n:بورنومو | 0.098% |
| 8 | economy | ← | n:√سهم n:عمان | 0.094% |
| 9 | economy | ← | v:√رجع n:√سهم | 0.081% |
| 10 | economy | ← | n:√بوس n:√سهم n:عمان | 0.076% |
| 11 | economy | ← | n:بيسة n:عمان | 0.076% |
| 12 | economy | ← | n:الاغلاقات | 0.076% |
| 13 | economy | ← | n:الاغلاقات n:√سهم | 0.062% |
| 14 | economy | ← | n:الاغلاقات n:بيسة n:√سهم | 0.060% |
| 15 | economy | ← | n:الاغلاقات n:بيسة | 0.060% |
| 16 | economy | ← | v:√رفع n:√سهم | 0.058% |
| 17 | religion | ← | v:√وصي n:√رجل | 0.058% |
| 18 | sports | ← | n:مانشستر | 0.058% |
| 19 | economy | ← | n:بورنومو v:√قلل | 0.054% |
| 20 | economy | ← | n:ريال n:√سهم | 0.051% |
| 21 | economy | ← | n:√سهم n:√سعر | 0.050% |
| 22 | economy | ← | n:بيسة v:√رفع n:√سهم | 0.048% |
| 23 | local | ← | v:√قبل n:√سعد | 0.048% |
| 24 | economy | ← | n:بيسة v:√رفع | 0.048% |
| 25 | sports | ← | n:مدريد n:ريال | 0.047% |
| 26 | sports | ← | n:ارسنال | 0.047% |
| 27 | sports | ← | n:يونايتد | 0.044% |
| 28 | economy | ← | n:بيسة n:√سهم n:√سعر | 0.043% |
| 29 | economy | ← | v:√رجع n:√سهم n:عمان | 0.043% |
| 30 | intern | ← | n:بوش v:√قلل | 0.043% |

The first column displays the rule consequent (the predicted class) while the third column displays the items of the rule, and the fourth column the rule's support (confidence is 100% for these rules).

The reader can observe the pertinence of these rules:

1) rules 1, 2, 5, 10, 11, 18, 19, 25, 26, 27, 30 contain proper nouns: OPEC, Purnomo Yusgiantoro (former president-secretary general of OPEC), Oman (the data of the *Kalimat* corpus originate from an Omani newspaper) for the Economy category, Manchester United, Arsenal and Real Madrid for the Sports category, Bush for the International news category;
2) many Economy rules contain the words ريال and بيسة (currency of Oman: 1 ريال = 1,000 بيسة), the roots √سهم (stock market share) and √سعر (rate);
3) Religion rules contain important words such as فتاوى (fatwa), احكام (provisions for reading the Koran, etc.), and roots such as √وصي (guardian), √رجل (man).

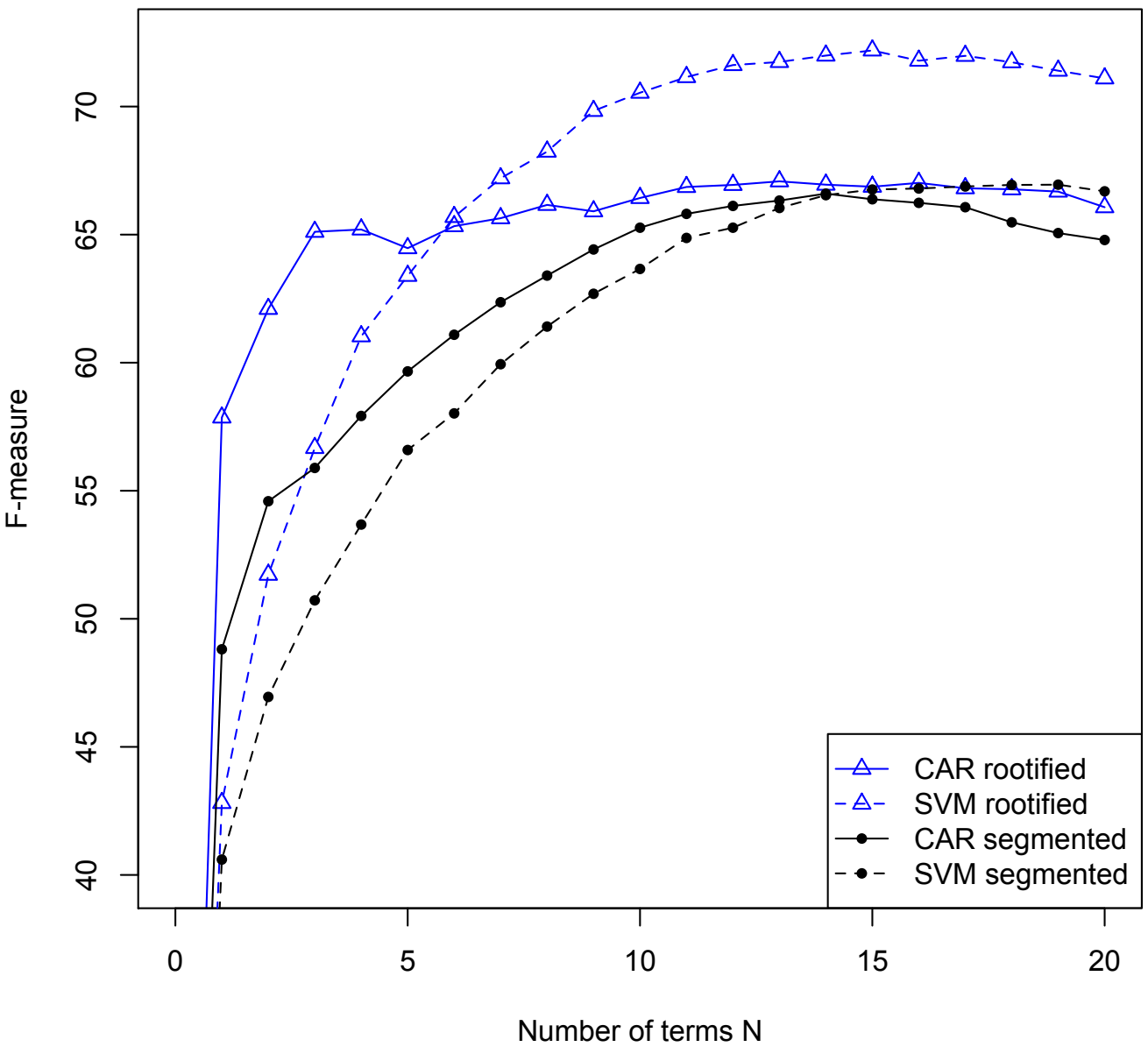

Fig. 2. F-measure as a function of the number of tfidf terms used for classification.

## V. Results

### A. Results Based on tfidf Feature Selection and Weighting

On Fig. 2 the reader can see the results (in terms of F-measure) of using CAR and SVM classifiers on features selected by tfidf. More precisely, we ranked (rootified or stemmed) words in each sentence by tfidf value and kept $N$ of them as features for the given sentence. The figure shows F-measure as a function of $N$, for values between 1 and 20. In the case of rootified data, the reader can observe that for small values of $N$ ($N \leq 5$) CARs give better results than SVM, that CARs stabilize around a F-measure of 67% and that SVMs rise until 72%. In the case of stemmed data, both CARs and SVMs give similar results and attain F-measures of 65–67%. At around $N = 20$ we have a small decrease of performance, probably an effect of semantic noise due to less relevant words.

### B. Results Based on Morphosyntactic Properties

In this section we investigate feature selection strategies based on morphosyntactic features. We will compare them with tfidf methods of similar transaction size.

Our approach will be the following: we will select features (either rootified or stemmed) by taking

1) only the head of the dependency tree (Strategy I);
2) only nouns neighboring the head (Strategy II);
3) the head plus nouns at distance 1 from the head (Strategy $\text{III}_1$);
4) the head plus nouns at distance $\leq$ 2 from the head (Strategy $\text{III}_2$);

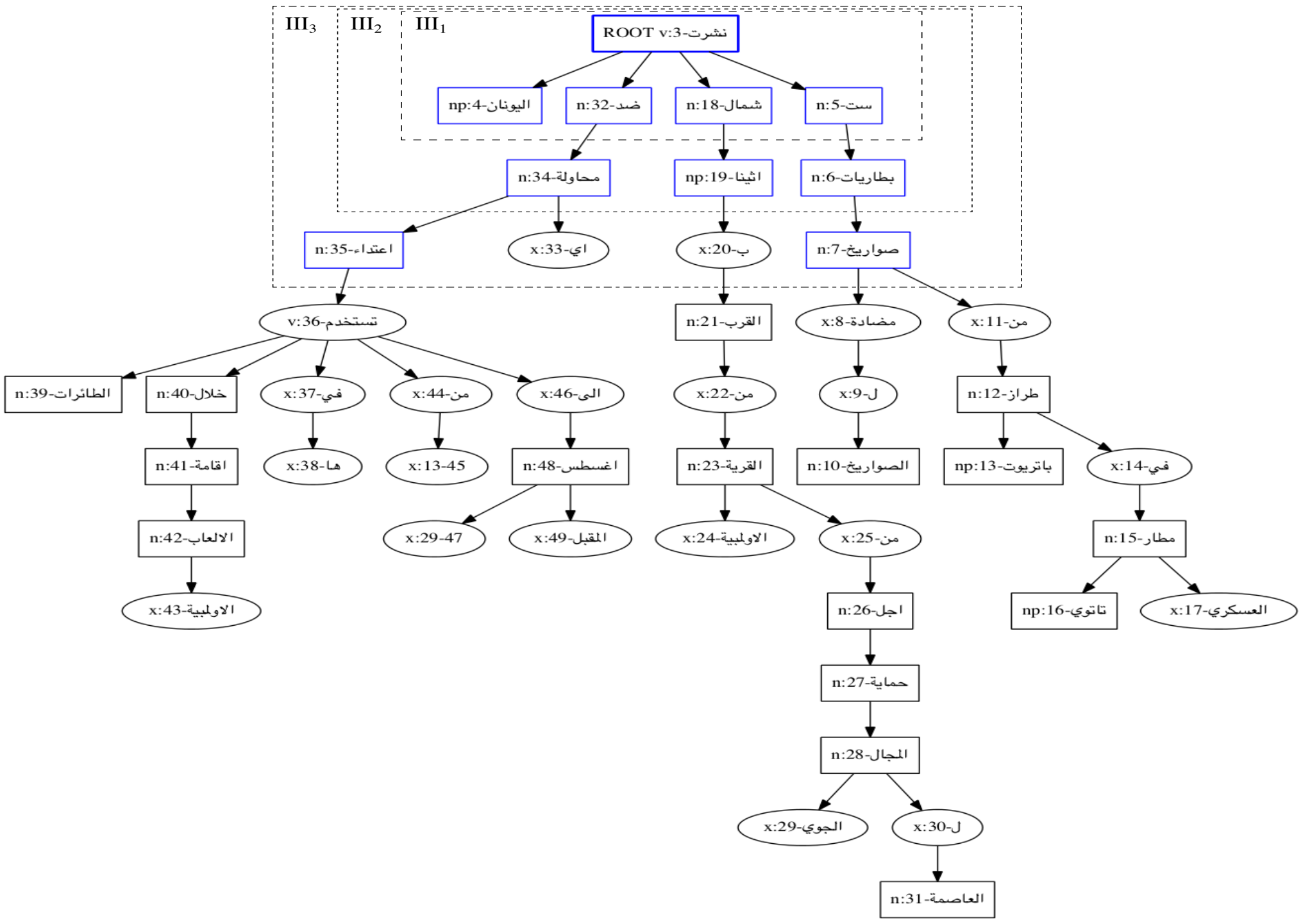

Fig. 3. Nodes selected by each strategy, for the example sentence.

5) the head plus nouns at distance $\leq 3$ from the head (Strategy $\text{III}_3$);
6) the head and all nouns of the sentence (Strategy IV);
7) the head, all nouns and all verbs of the sentence (Strategy IV′).

The reader can see an illustration of the application of these strategies to the example sentence of the previous section, in Fig. 3: boxes are nouns, blue boxes are those taken by Strategies II–$\text{III}_3$, dashed boxes show the limits of strategies. Strategy IV will consider all nouns, IV′ will consider all nouns and verbs of the sentence.

*1) Strategy I:* Our first strategy will be to keep only the head (in the sense of dependency grammar) of each sentence, provided it is a noun (which happens in 5.12% of cases), proper noun (in 2.15% of cases), or verb (in 81.1% of cases).

On Fig. 4 the reader can see the results obtained by applying this strategy to rootified (above) and stemmed (below) data.

| | CUL | LOC | INT | SPO | REL | ECO | AVG |
|---|---|---|---|---|---|---|---|
| CAR Recall | 53.33 | 37.20 | 86.60 | 49.38 | 58.61 | 35.90 | 45.86 |
| CAR Precision | 48.43 | 63.89 | 51.91 | 79.47 | 73.01 | 68.04 | 54.96 |
| CAR F-measure | 50.76 | 47.02 | 64.91 | 60.91 | 65.02 | 47.00 | 47.95 |
| MinSupp=0.001, MinConf=49.2, AvgTransSize=1.00 | | | | | | | |
| SVM Recall | 45.04 | 19.12 | 55.16 | 25.21 | 30.68 | 20.51 | 32.62 |
| SVM Precision | 35.79 | 22.75 | 37.21 | 33.38 | 38.75 | 33.33 | 33.53 |
| SVM F-measure | 39.88 | 20.78 | 44.44 | 28.73 | 34.24 | 25.40 | 32.24 |

| | CUL | LOC | INT | SPO | REL | ECO | AVG |
|---|---|---|---|---|---|---|---|
| CAR Recall | 58.34 | 41.38 | 83.39 | 57.10 | 59.09 | 36.49 | 47.97 |
| CAR Precision | 51.59 | 62.69 | 57.52 | 80.08 | 76.74 | 68.39 | 56.71 |
| CAR F-measure | 54.75 | 49.85 | 68.08 | 66.66 | 66.76 | 47.59 | 50.53 |
| MinSupp=0.001, MinConf=48.4, AvgTransSize=1.00 | | | | | | | |
| SVM Recall | 45.50 | 26.97 | 50.16 | 31.05 | 28.35 | 29.04 | 35.18 |
| SVM Precision | 33.15 | 33.72 | 45.28 | 37.53 | 49.19 | 26.49 | 37.56 |
| SVM F-measure | 38.35 | 29.97 | 47.59 | 33.99 | 35.97 | 27.71 | 35.60 |

Fig. 4. Results of Strategy I applied to rootified (above) and stemme (below) data.

Globally these results are rather bad and show that the head alone is not a good feature to select. The results obtained from stemmed data are slightly better than those from rootified data. There are some interesting outliers: for example, the recall of International news is quite high (in the case of the CAR classifier), they are probably easy to detect because of the many foreign proper nouns. Also very high is the precision of Sports (again in the case of the CAR classifier only), and since heads are mostly verbs, we can deduce from this result that sport verbs allow identification of sport text with high precision. The same phenomenon (in a slightly lesser degree) can be observed for Religion. In these three cases there is an enormous difference between the performance of CARs and SVMs.

The rules generated in this strategy have a transaction size of one (!), in other words there are of the form $i \to c$ where $i$ is a single item and $c$ is the consequent (the predicted class). This is an important issue because CPU charge increases exponentially with respect to transaction size.

Before we move to the next strategy, let us compare performances of Strategy I and of tfidf feature selection. To be fair, we must compare performance while keeping the same transaction size. For tfidf, a transaction size of 1 is obtained for $N = 1$, and the F-measure in this case is of 57.86% for rootified data and of 48.81% for stemmed data. It follows that Strategy I is slightly better than tfidf feature selection for stemmed data, but significantly less effective (almost 10% less F-measure) in the case of rootified data.

*2) Strategy II:* Since heads are not sufficiently discriminative features, we will now turn to the neighboring nodes in the dependency grammar tree: let us take nodes located at distance 1 from the head. In order to increase pertinence and reduce size of our itemsets, we will restrict ourselves to nouns and proper nouns.

On Fig. 5 the reader can see the results obtained by applying this strategy to rootified (above) and stemmed (below) data.

| | CUL | LOC | INT | SPO | REL | ECO | AVG |
|---|---|---|---|---|---|---|---|
| CAR Recall | 66.39 | 52.99 | 87.81 | 75.26 | 70.53 | 48.49 | 57.35 |
| CAR Precision | 70.28 | 66.08 | 67.17 | 86.08 | 81.04 | 78.13 | 64.11 |
| CAR F-measure | 68.28 | 58.82 | 76.12 | 80.31 | 75.42 | 59.84 | 59.83 |
| MinSupp=0.004, MinConf=55, AvgTransSize=1.63 | | | | | | | |
| SVM Recall | 52.73 | 40.03 | 55.32 | 55.15 | 43.20 | 34.45 | 46.81 |
| SVM Precision | 44.25 | 37.97 | 50.93 | 48.22 | 58.25 | 43.97 | 47.27 |
| SVM F-measure | 48.12 | 38.97 | 53.04 | 51.45 | 49.61 | 38.63 | 46.64 |

| | CUL | LOC | INT | SPO | REL | ECO | AVG |
|---|---|---|---|---|---|---|---|
| CAR Recall | 73.22 | 64.20 | 88.91 | 77.83 | 74.72 | 58.92 | 62.54 |
| CAR Precision | 69.22 | 66.71 | 72.16 | 90.83 | 84.35 | 78.57 | 65.98 |
| CAR F-measure | 71.17 | 65.43 | 79.67 | 83.82 | 79.24 | 67.34 | 63.81 |
| MinSupp=0.004, MinConf=46.2, AvgTransSize=1.64 | | | | | | | |
| SVM Recall | 57.96 | 44.98 | 61.42 | 53.92 | 52.30 | 44.40 | 52.50 |
| SVM Precision | 49.56 | 43.25 | 57.15 | 58.75 | 60.67 | 53.02 | 53.73 |
| SVM F-measure | 53.43 | 44.10 | 59.21 | 56.23 | 56.18 | 48.33 | 52.91 |

Fig. 5. Results of Strategy II applied to rootified (above) and stemme (below) data.

Surprisingly, the results are *much* better than those of Strategy I: we observe a performance increase of 29.6% for CARs and of 51.3% for SVMs (rootified data). This is due to two factors: the semantic pertinence of nouns dependent on the head, and the increase of the average transaction size from 1 (Strategy I) to 2.25 (Strategy II).

Otherwise we still observe the small difference between rootified and stemmed data as well as the outliers we encountered in Strategy I.

Let us now compare this strategy with tfidf feature selection: in the case of rootified data, we get a comparable average transaction size for $N = 2$; in that case F-measure is 62.10, which is almost exactly the result we get in Strategy II. In the case of stemmed data, we get a comparable average transaction size for $N = 3$, that corresponds to a F-measure of 55.89, which is, once again, significantly less efficient than Strategy II.

*3) Strategy $III_1$:* Our next attempt will be to combine the head node with neighboring noun (and proper noun) nodes. This approach will be incrementally generalized in Strategies $III_2$ and $III_3$: in fact we take, in Strategy $III_n$, all nouns and proper nouns at distance $\leq n$. On Fig. 6 the reader can see the results obtained by applying strategy $III_1$ to rootified (above) and stemmed (below) data ($n = 1$).

| | CUL | LOC | INT | SPO | REL | ECO | AVG |
|---|---|---|---|---|---|---|---|
| CAR Recall | 71.05 | 61.14 | 91.72 | 75.63 | 74.04 | 51.69 | 60.75 |
| CAR Precision | 70.54 | 68.67 | 66.29 | 88.75 | 83.91 | 77.55 | 65.10 |
| CAR F-measure | 70.80 | 64.69 | 76.96 | 81.66 | 78.67 | 62.04 | 62.12 |
| MinSupp=0.004, MinConf=55, AvgTransSize=2.25 | | | | | | | |
| SVM Recall | 56.22 | 35.30 | 64.44 | 47.12 | 50.79 | 42.67 | 49.42 |
| SVM Precision | 59.82 | 37.88 | 59.74 | 51.16 | 34.82 | 49.41 | 48.81 |
| SVM F-measure | 57.97 | 36.54 | 62.00 | 49.05 | 41.32 | 45.79 | 48.78 |

| | CUL | LOC | INT | SPO | REL | ECO | AVG |
|---|---|---|---|---|---|---|---|
| CAR Recall | 76.10 | 68.67 | 93.59 | 78.21 | 80.39 | 56.06 | 64.72 |
| CAR Precision | 71.95 | 68.68 | 69.60 | 90.82 | 88.13 | 79.49 | 66.95 |
| CAR F-measure | 73.96 | 68.67 | 79.84 | 84.04 | 84.08 | 65.75 | 65.19 |
| MinSupp=0.004, MinConf=47.4, AvgTransSize=2.25 | | | | | | | |
| SVM Recall | 53.59 | 39.14 | 65.76 | 51.65 | 52.46 | 36.76 | 49.89 |
| SVM Precision | 48.13 | 40.44 | 54.19 | 66.14 | 48.12 | 50.14 | 51.19 |
| SVM F-measure | 50.71 | 39.78 | 59.42 | 58.01 | 50.20 | 42.42 | 50.09 |

Fig. 6. Results of Strategy $III_1$ applied to rootified (above) and stemme (below) data ($n = 1$).

As we can see, the results are slightly better (by about 3%) than Strategy II, while the average transaction size is still low: 2.25. The high recall of International News and the high precision of Sports and Religion have increased as well. The results of the SVM classifier are still around 15% behind those of the CAR classifier. Finally, to compare with tfidf feature selection we need to consider tfidf results based on the same transaction size: for rootified text this is the case for $N = 2$, the F-measure of which is 62.10% which is equal to the $III_2$ result; for stemmed text, $N$ will rather be 3, the F-measure of which is 55.89%, which is way beneath the corresponding $III_2$ result.

To conclude: $III_2$ gives better results than II, without increasing average transaction size beyond measure, CAR classifier is better than SVM, and dependency grammar equal in performance to tfidf for rootified text and better than tfidf for stemmed text.

*4) Strategy $III_2$:* In this case we take all nouns and proper nouns at distance 1 or 2 from the head. On Fig. 7 (p. 8) the reader can see the results obtained by applying this strategy to rootified (above) and stemmed (below) data.

We observe a significant improvement compared to Strategy $III_1$ but also that the average transaction size has almost doubled: it is now around 4.28. Recall of International News and precision of Religion have increased, although not as much as the global F-measure, while precision of Sports has not changed. SVMs still perform less than CARs, but the difference between the two has decreased.

| | CUL | LOC | INT | SPO | REL | ECO | AVG |
|---|---|---|---|---|---|---|---|
| CAR Recall | 72.44 | 77.49 | 93.17 | 86.26 | 81.40 | 57.44 | 66.89 |
| CAR Precision | 80.17 | 66.34 | 76.25 | 88.75 | 87.84 | 82.14 | 68.79 |
| CAR F-measure | 76.11 | 71.49 | 83.87 | 87.49 | 84.50 | 67.61 | 67.29 |

MinSupp=0.015, MinConf=52.6, AvgTransSize=4.28

| | | | | | | | |
|---|---|---|---|---|---|---|---|
| SVM Recall | 53.56 | 51.25 | 72.92 | 66.78 | 64.70 | 46.50 | 59.28 |
| SVM Precision | 69.47 | 54.67 | 54.02 | 66.78 | 60.43 | 49.33 | 59.12 |
| SVM F-measure | 60.49 | 52.90 | 62.06 | 66.78 | 62.49 | 47.87 | 58.77 |

| | CUL | LOC | INT | SPO | REL | ECO | AVG |
|---|---|---|---|---|---|---|---|
| CAR Recall | 74.58 | 76.63 | 95.16 | 87.18 | 85.30 | 63.98 | 68.97 |
| CAR Precision | 83.80 | 69.63 | 77.00 | 92.81 | 88.88 | 82.68 | 70.69 |
| CAR F-measure | 78.92 | 72.96 | 85.12 | 89.91 | 87.06 | 72.13 | 69.44 |

MinSupp=0.015, MinConf=54.2, AvgTransSize=4.29

| | | | | | | | |
|---|---|---|---|---|---|---|---|
| SVM Recall | 56.91 | 45.32 | 78.65 | 65.78 | 56.07 | 49.68 | 58.73 |
| SVM Precision | 62.38 | 56.47 | 54.74 | 67.29 | 64.10 | 53.44 | 59.74 |
| SVM F-measure | 59.52 | 50.29 | 64.56 | 66.52 | 59.82 | 51.49 | 58.70 |

Fig. 7. Results of Strategy $III_2$ applied to rootified (above) and stemme (below) data.

To compare with tfidf feature selection, we need to take $N = 4$ for rootified data and $N = 7$ for stemmed data. In the first case, the F-measure is 65.20%, which is a bit less than $III_2$, and in the second case, F-measure is 62.36% which is significantly worse than $III_2$.

*5) Strategy $III_3$:* In this case we take all nouns and proper nouns at distance 1, 2 or 3 from the head. On Fig. 8 the reader can see the results obtained by applying this strategy to rootified (above) and stemmed (below) data.

| | CUL | LOC | INT | SPO | REL | ECO | AVG |
|---|---|---|---|---|---|---|---|
| CAR Recall | 72.02 | 86.08 | 94.58 | 88.30 | 89.04 | 63.42 | 70.49 |
| CAR Precision | 87.82 | 65.31 | 81.52 | 94.55 | 89.71 | 83.65 | 71.79 |
| CAR F-measure | 79.14 | 74.27 | 87.57 | 91.32 | 89.38 | 72.14 | 70.54 |

MinSupp=0.022, MinConf=46.6, AvgTransSize=6.44

| | | | | | | | |
|---|---|---|---|---|---|---|---|
| SVM Recall | 54.23 | 53.96 | 76.92 | 75.91 | 67.00 | 57.08 | 64.18 |
| SVM Precision | 68.96 | 56.54 | 58.31 | 76.55 | 58.98 | 65.12 | 64.08 |
| SVM F-measure | 60.71 | 55.22 | 66.33 | 76.23 | 62.73 | 60.83 | 63.68 |

| | CUL | LOC | INT | SPO | REL | ECO | AVG |
|---|---|---|---|---|---|---|---|
| CAR Recall | 71.57 | 84.49 | 95.89 | 91.10 | 90.63 | 68.46 | 71.73 |
| CAR Precision | 90.65 | 67.97 | 85.62 | 95.39 | 88.59 | 84.06 | 73.18 |
| CAR F-measure | 79.99 | 75.33 | 90.46 | 93.20 | 89.60 | 75.46 | 72.01 |

MinSupp=0.018, MinConf=58.6, AvgTransSize=6.47

| | | | | | | | |
|---|---|---|---|---|---|---|---|
| SVM Recall | 59.85 | 58.74 | 78.87 | 79.48 | 69.49 | 63.33 | 68.29 |
| SVM Precision | 77.42 | 61.70 | 62.29 | 74.16 | 60.37 | 65.14 | 66.85 |
| SVM F-measure | 67.51 | 60.19 | 69.61 | 76.73 | 64.61 | 64.22 | 67.14 |

Fig. 8. Results of Strategy $III_3$ applied to rootified (above) and stemme (below) data.

Performance has still increased in quite homogeneous way, by around 3%. The average transaction size (6.44) is higher than for $III_2$, but remains in reasonable bounds. Recall of International News and precision of Sports seem to have reached their highest possible values which are around 95–96%.

To compare with tfidf feature selection, we need to take $N = 7$ for rootified data and $N = 13$ for stemmed data. In the first case, the F-measure is 65.64%, and in the second case, F-measure is 66.33%. Both of them are significantly worse than $III_3$.

*6) Strategies IV and IV′:* Instead of continuing in the same realm and defining $III_n$ for $n \geq 4$, we will adopt a new strategy and use *all* nouns of each sentence. The goal is to show that this new strategy (called Strategy IV) has only slightly better performance than $III_3$, while the average transaction size grows significantly.

On Fig. 9 the reader can see the results obtained by applying this strategy to rootified (above) and stemmed (below) data.

| | CUL | LOC | INT | SPO | REL | ECO | AVG |
|---|---|---|---|---|---|---|---|
| CAR Recall | 61.88 | 86.21 | 95.10 | 92.91 | 94.27 | 77.08 | 72.49 |
| CAR Precision | 96.12 | 70.89 | 89.75 | 91.55 | 83.98 | 83.94 | 73.75 |
| CAR F-measure | 75.29 | 77.80 | 92.35 | 92.23 | 88.83 | 80.36 | 72.41 |

MinSupp=0.051, MinConf=67.6, AvgTransSize=13.76

| | | | | | | | |
|---|---|---|---|---|---|---|---|
| SVM Recall | 60.63 | 55.60 | 89.94 | 86.14 | 72.12 | 79.84 | 74.05 |
| SVM Precision | 81.70 | 66.77 | 75.04 | 86.33 | 73.65 | 70.15 | 75.61 |
| SVM F-measure | 69.61 | 60.67 | 81.82 | 86.23 | 72.88 | 74.68 | 74.32 |

| | CUL | LOC | INT | SPO | REL | ECO | AVG |
|---|---|---|---|---|---|---|---|
| CAR Recall | 63.71 | 86.50 | 95.96 | 93.50 | 94.26 | 79.32 | 73.32 |
| CAR Precision | 96.17 | 71.74 | 89.18 | 95.55 | 85.02 | 84.07 | 74.53 |
| CAR F-measure | 76.64 | 78.43 | 92.45 | 94.51 | 89.40 | 81.63 | 73.29 |

MinSupp=0.052, MinConf=68, AvgTransSize=13.90

| | | | | | | | |
|---|---|---|---|---|---|---|---|
| SVM Recall | 62.88 | 68.19 | 89.77 | 86.49 | 80.36 | 82.29 | 78.33 |
| SVM Precision | 86.22 | 57.10 | 76.41 | 86.92 | 80.79 | 78.62 | 77.68 |
| SVM F-measure | 72.72 | 62.15 | 82.55 | 86.71 | 80.57 | 80.41 | 77.52 |

Fig. 9. Results of Strategy IV applied to rootified (above) and stemme (below) data.

As expected, the results are better than those of $III_3$ but *only* by 1.87% for rootified data and by 1.28% for stemmed data (!). On the other hand, average transaction sizes are more than the double than the corresponding $III_3$: 13.76 for rootified and 13.90 for stemmed text. This shows clearly that it not necessary to go beyond $III_3$, if CARs are to be used.

Nevertheless, the spectacular result of this strategy is the performance of SVMs: not only has it increased, but it now exceeds significantly the one of CARs. This shows that SVMs become interesting classifiers when the average transaction size gets high.

How about going even farther, and taking also *all verbs* of the sentence? We have tried this strategy (which we call IV′), and the results are disappointing. On Fig. 10 (p. 9) the reader can see the results obtained by applying Strategy IV′ to rootified (above) and stemmed (below) data.

As the reader can see, while average transaction size has increased as expected, performance has decreased, and this goes both for CARs and for SVMs.

| | CUL | LOC | INT | SPO | REL | ECO | AVG |
|---|---|---|---|---|---|---|---|
| CAR Recall | 58.48 | 85.48 | 93.45 | 92.35 | 95.53 | 75.56 | 71.55 |
| CAR Precision | 97.08 | 70.14 | 90.22 | 91.76 | 77.87 | 84.19 | 73.04 |
| CAR F-measure | 72.99 | 77.05 | 91.81 | 92.05 | 85.80 | 79.64 | 71.33 |
| MinSupp=0.069, MinConf=64, AvgTransSize=16.23 | | | | | | | |
| SVM Recall | 57.01 | 56.37 | 90.97 | 77.76 | 74.47 | 77.34 | 72.32 |
| SVM Precision | 87.18 | 56.85 | 64.53 | 80.81 | 72.78 | 69.37 | 71.92 |
| SVM F-measure | 68.94 | 56.61 | 75.50 | 79.26 | 73.61 | 73.14 | 71.18 |

| | CUL | LOC | INT | SPO | REL | ECO | AVG |
|---|---|---|---|---|---|---|---|
| CAR Recall | 60.66 | 87.06 | 95.66 | 92.46 | 96.15 | 75.75 | 72.53 |
| CAR Precision | 97.49 | 68.52 | 88.93 | 95.91 | 82.26 | 84.14 | 73.89 |
| CAR F-measure | 74.79 | 76.68 | 92.18 | 94.15 | 88.66 | 79.72 | 72.31 |
| MinSupp=0.073, MinConf=59.2, AvgTransSize=16.42 | | | | | | | |
| SVM Recall | 64.27 | 62.59 | 87.80 | 82.33 | 70.49 | 80.36 | 74.64 |
| SVM Precision | 82.98 | 67.09 | 65.93 | 85.83 | 73.00 | 73.57 | 74.73 |
| SVM F-measure | 72.44 | 64.76 | 75.31 | 84.04 | 71.72 | 76.81 | 74.18 |

Fig. 10. Results of Strategy IV′ applied to rootified (above) and stemme (below) data.

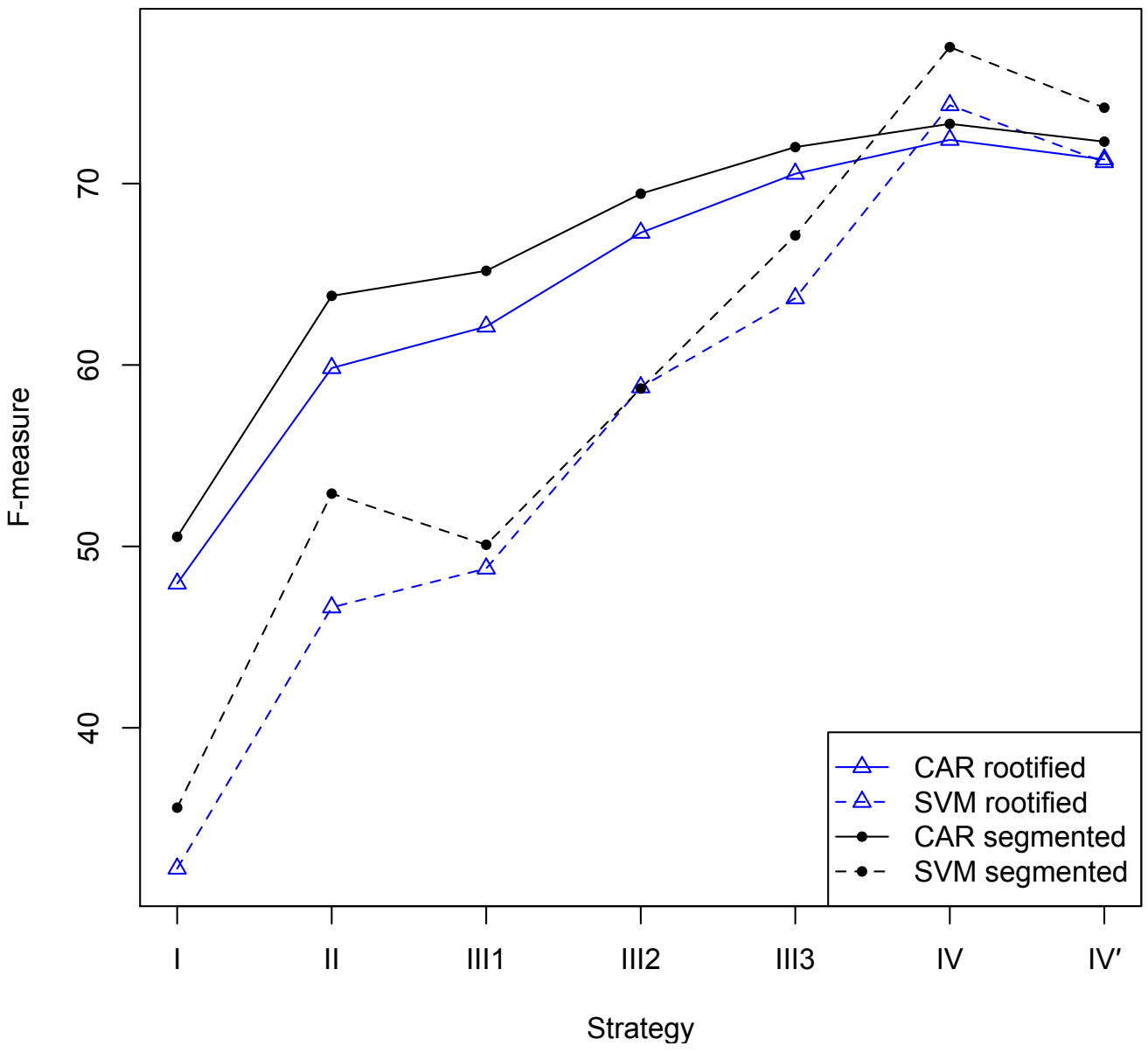

Fig. 11. F-measure achieved by each strategy.

*7) Synthesis:* We are synthesizing the results of this section in Fig. 11, where we represent F-measure as a function of the different strategies we used. One can see clearly that stemmed text gives always slighly better results than rootified text, and also that SVMs are significantly less powerful in small transaction size strategies, but have a higher slope and hence exceed CAR strategies in Strategy IV.

## VI. Conclusion

In this paper we have studied various text classification techniques applied to Arabic text. We have prepared two versions of the corpus: one with rootified words and one with stemmed words. We have considered sentences as transactions of features which we have selected in two ways: by tfidf techniques and by dependency grammar techniques. Finally we have applied two kinds of classifiers to our data: class association rules and support vector machines.

The results can be synthesized as follows:

1) classification of stemmed text is slightly better than classification of rootified text[2];
2) dependency grammar feature selection gives better results when compared over transactions of the same average size;
3) class association rules are better than support vector machines in small transaction sizes but get better when transaction size gets high (in which case dependency grammar is not used anymore and the feature selection criteria are purely morphological).

[2]It would be interesting to do the same calculations over lemmatized words, rather than rootified and stemmed ones.

Furthermore, the results we presented can be improved since *Kalimat*'s word/root correspondence has a non-negligible error ratio of 7.83%, more precisely:

| Category | CUL | LOC | INT | SPO | REL | ECO | AVG |
|---|---|---|---|---|---|---|---|
| Error in % | 11.24 | 8.42 | 8.50 | 4.84 | 4.28 | 9.7 | 7.83 |

(These results have been obtained by manual inspection of randomly chosen sets of 10 texts per category.) We expect to get better results for rootified text, when applied to a corrected version of *Kalimat*.